\documentclass[aps,pra,twocolumn,amsmath,amssymb,superscriptaddress,reprint,longbibliography]{revtex4-2}

\usepackage[utf8]{inputenc}
\usepackage{graphicx}
\usepackage[colorlinks=true,citecolor=blue]{hyperref}
\usepackage[caption=false]{subfig}
\usepackage{lipsum}
\usepackage{float}

\begin{document}

\title{Entropy and type-token ratio in gigaword corpora}

\author{Pablo Rosillo-Rodes}
\email{prosillo@ifisc.uib-csic.es}
\author{Maxi San Miguel}
\author{David Sánchez}

\affiliation{
 Institute for Cross-Disciplinary Physics and Complex Systems IFISC (UIB-CSIC), Campus Universitat de les Illes Balears, E-07122 Palma de Mallorca, Spain
}
\date{\today}

\begin{abstract}
There are different ways of measuring diversity in complex systems. In particular, in language, lexical diversity is characterized in terms of the type-token ratio and the word entropy. We here investigate both diversity metrics in six massive linguistic datasets in English, Spanish, and Turkish, consisting of books, news articles, and tweets. These gigaword corpora correspond to languages with distinct morphological features and differ in registers and genres, thus constituting a varied testbed for a quantitative approach to lexical diversity. We unveil an empirical functional relation between entropy and type-token ratio of texts of a given corpus and language, which is a consequence of the statistical laws observed in natural language. Further, in the limit of large text lengths we find an analytical expression for this relation relying on both Zipf and Heaps laws that agrees with our empirical findings. 
\end{abstract}

\maketitle

\section{Introduction}

Type-token ratio ($TTR$) and entropy ($H$) are useful metrics for evaluating the lexical diversity or richness. $TTR$ is defined as
\begin{equation}
    \label{eq:ttr}
    TTR = \frac{V}{L},
\end{equation}
and simply measures the proportion of unique words (types), i.e., the vocabulary size, $V$, to the total number of words (tokens) in a given text, $L$~\cite{Johnson1944,Templin1957, hardie2006, Kettunen2014, Bermel2015, litvinova2017}. $TTR$ has been originally leveraged in the study of children vocabulary~\cite{Templin1957, hess1986} or illnesses which affect linguistic capabilities~\cite{Manschreck1984}. Diversity measures, however, are not only interesting for linguistics. For example, in ecology, the Hill number of order 0 measures the total number of different species (or types) in a population of a given size~\cite{Jost2006, Tuomisto2010, Mazzarisi2021}.

In contrast, entropy takes into account the distribution of the unique words in terms of their occurrence frequencies, capturing both the diversity of lexical items and also how uniformly they are used. The Shannon entropy is defined as
\begin{equation}
\label{eq:shannonh}
    H = -\sum\limits_{i=1}^V p_i \ln p_i,
\end{equation}
where $p_i$ represents the probability of the $i$-th word within vocabulary $V$. As such, understanding the word entropy of texts~\cite{Bentz2017} offers valuable insights into the complexity and unpredictability of language~\cite{montemurro2011universal}.  Originally proposed within information theory~\cite{shannon1948,cover1999}, in the context of language higher entropy suggests more unpredictability in word choices, which can reveal underlying structures of language use and cognitive processing. Texts with high entropy exhibit greater diversity in word usage, reflecting more complex linguistic structures. Specifically, this measure and related operationalizations have been applied in a broad range of linguistic studies, including differentiation between translated and original documents~\cite{liu2022}, the study of ambiguity in different legal systems~\cite{friedrich2020}, quantification of language change~\cite{Gerlach2016}, identification of stopwords~\cite{gerlach2019stopwords}, or observation of change in literary preference~\cite{Mohseni2023}, among others. In the context of ecology, entropy is used as a measure of diversity in studies about populations~\cite{Eliazar2010} and mutations in tumoral cells~\cite{Romeroarias2018}, among others, under the name of Shannon index. Indeed, the Hill number of order 1 is simply $\exp(H)$, similar to perplexity~\cite{Jost2006, Tuomisto2010}.

Hence, type-token ratio and entropy are two widely-used metrics that address diversity measurement in different ways. For example, consider a language, $\mathcal{L}$, consisting of only two words, $g$ and $m$. The vocabulary of this language can be expressed as $V_\mathcal{L} = \{ g, m \}$. One might initially assume that it is possible to create multiple texts of the same length and type-token ratio while still achieving different entropy values. We show this in Table~\ref{tab:test_text}, where different strings with the same $TTR$ yield different values of the entropy because of their different probability distributions. This observation would suggest that type-token ratio and entropy are distinct and uncorrelated diversity measures.

\begin{table}[t] 
\caption{Three possible samples of texts of $L=10$ for a language having a vocabulary size $V=2$. Note that the samples exhibit the same $TTR$ but different entropy.\label{tab:test_text}}
\begin{tabular}{ccccc}
\hline \hline
Text & $TTR$ & $p_1$ & $p_2$ & $H$ (nats) \\
\hline
$g m m g m g m g g m$ & $1/5$ & $1/2$ & $1/2$ & 0.69   \\
$g m m m g m m m m m$ & $1/5$ & $1/5$ & $4/5$ & 0.50 \\
$g m m m m m m m m m$ & $1/5$ & $1/10$ & $9/10$ & 0.33    \\
\hline \hline
\end{tabular}
\end{table}

Nonetheless, the production of natural language texts is constrained by statistical laws. These empirical rules highlight patterns that occur in word frequency and vocabulary growth across languages. On the one hand, Zipf law~\cite{zipf1946} states that the distribution of word frequencies in a text follows a power law. On the other hand,  Heaps law~\cite{herdan1960,Heaps1978} dictates that the number of types in a text increases sublinearly with its total number of tokens. These laws, among others, are found in numerous linguistic corpora~\cite{Altmann2016, Stanisz2024, Arnon2025} and help linguists and computational scientists model language, offering insights into communication efficiency and the structure of human language systems. Their persistence among different multilingual corpora is rooted in the balance between repetition and novelty in human communication: a few common words are used very frequently for efficiency, while rarer words carry specificity and keep appearing as texts grow, though at a decreasing rate. Several generative models have been proposed to explain Zipf and Heaps laws and other statistical features of natural language, often highlighting their links to cognitive processes~\cite{Mitzenmacher2004,Serrano2009,Gerlach2013}.

Quite generally, Zipf and Heaps laws are closely related to the concept of diversity in text analysis. Zipf law, by highlighting the unequal distribution of word frequencies, influences the predictability of words in a text. As mentioned above, this distribution plays a key role in calculating the entropy of natural language texts, where frequent words reduce unpredictability and rarer words increase it. Similarly, Heaps law, which describes vocabulary growth, directly affects the lexical diversity of a text as calculated from the $TTR$ definition given by Eq.~\eqref{eq:ttr}. 

Type-token ratio and entropy are known to highly depend on sample size~\cite{Malvern2004, Shi2012, Gregori2015, Koplenig2019, Magurran2021}. This dependence is indirectly due to the statistical laws of natural language discussed above. From a dependence of both $TTR$ and $H$ on $L$ as independent variable, it follows that, contrary to what is deduced from Table~\ref{tab:test_text}, the value of $H$ should depend on the value of $TTR$. In this work, we explore this possible functional relation between type-token ratio and entropy in large corpora.

Our empirical findings are based on the analysis of six gigaword corpora from different sources and languages. We further verify that Zipf and Heaps laws are observed in our corpora with over one billion words, as expected. Based on this analysis, we present two major findings.

First, we identify an empirical functional relation between the entropy and the type-token ratio. This relation is consistently found across the studied massive corpora. Second, to gain more insight, we fit an analytical expression for the word entropy to the corpus data. This expression is based on the validity of statistical laws of natural language and holds for large vocabularies. Using Heaps law, we express the word entropy in terms of the type-token ratio and fit it to the six massive corpora, finding an excellent agreement with the numerical results for large texts. The agreement is observed for languages of different morphological types and sources of distinct genres and registers or styles. This is particularly relevant since it suggests that the $H$-$TTR$ relation is robust against how language users vary their word choices, at least in large corpora. Our findings would then be valuable in optimizing performance and efficiency across various tasks and applications of natural language processing (NLP) and machine learning including quantifying information content, evaluating large language models and enhancing data compression. 

The remaining of the manuscript is structured as follows. In Section~\ref{sec:data}, we describe the corpora used and analyze their corresponding Zipf and Heaps laws. In Section~\ref{sec:entropy} we study empirical measurements of the entropies in our gigaword corpora. We further derive an analytical expression for the entropy of large texts and fit this formula to the different datasets, obtaining good results. In Section~\ref{sec:ttr} we investigate the type-token ratio in the various corpora, both empirically and analytically using the Heaps law. We combine both diversity metrics in Section~\ref{sec:entropyandttr}, where we discuss the functional relation that connects $H$ and $TTR$ and its agreement with the empirical results. Finally, Section~\ref{sec:conclusions} contains our conclusions and suggestions for further work.

\section{Data}
\label{sec:data}

We utilize six large-scale text corpora, each containing over $10^9$ words. These corpora are drawn from three primary sources: books, online news articles, and Twitter (now X). The languages represented in these corpora are English, Spanish, and Turkish, selected to reflect varying degrees of isolating, inflectional and agglutinative morphology of their vocabularies. This is important since lexical items are differently formed in the three languages. English, despite being a fusional language, shows features of isolating languages, where each word typically consists of a single morpheme. Spanish comparatively uses more inflectional structures that convey multiple morphosyntactic traits. Finally, Turkish is considered as an agglutinative language with a high number of morphemes per word. Therefore, these languages lie in different positions within the morphological classification continuum~\cite{comrie1989language}. Obviously, this selection is not comprehensive but is representative and sufficient for the purposes of our work.

Since there exist subtleties in the definition of word~\cite{taylor2015oxford}, we hereafter apply a pragmatic approach and consider word as every linguistic unit that appears in our corpora between spaces or punctuation marks. This orthographic identification facilitates word counting and is the preferred criterion in commonly used NLP tokenizers~\cite{bird2006nltk}.

The preprocessing criteria of the texts forming the corpora follow those established by Gerlach et al.~\cite{Gerlach2020}, including tokenization, removal of punctuation and numbers, conversion to lowercase, and preservation of accents. The corpora which are publicly available may be found in Ref.~\cite{repo}. Below is a brief overview, as summarized in Table~\ref{tab:corpora}.

The first corpus, the Standardized Project Gutenberg Corpus (SPGC), introduced by Gerlach et al.~\cite{Gerlach2020}, consists of over $5\times10^4$ curated books from Project Gutenberg~\cite{PG}. SPGC has been widely used in quantitative linguistics research~\cite{dunn2022}, from the study of changes in the use of vernacular biological names~\cite{langer2021} and understanding large-language models~\cite{he2022}, to measuring the similarity of texts~\cite{Shade2023} and studying statistical laws in natural language~\cite{Altmann2016, Corral2020}. For our analysis, we select books from the SPGC written in English, excluding dictionaries and glossaries, resulting in a sample of 41,418 books. This subset contains 2,620,190,214 tokens and 3,144,125 types, of which 1,327,534 (40\%) are hapax legomena (words with only one occurrence in the corpus).

The second corpus (SPA) is part of the Corpus del Español~\cite{Davies2016}, constituting a large collection of texts in Spanish from web pages of 21 Spanish-speaking countries, gathered between 2013 and 2014. After preprocessing, this corpus comprises 1,887,720,265 tokens and 3,699,411 types, of which 1,900,485 (51\%) are hapax legomena. The Corpus del Español is
widely used in quantitative linguistics of Spanish, from multi-dimensional analysis of register variation~\cite{Biber2006} to comparisons between corpora and live expositions to certain language forms~\cite{Daidone2019}.

The third corpus, the Turkish subcorpus of CC100 (TRCC100), is a collection of web crawl data from Common Crawl~\cite{commoncrawl} in Turkish compiled by Facebook AI team using CCNet~\cite{conneau2020, fbaiccnet}. After preprocessing, this corpus contains 2,640,929,269 tokens and 9,804,688 types, and 5,159,681 hapax legomena (53\% of the types).

The following three corpora are compilations of geolocalized posts from the microblogging website Twitter (now X)~\footnote{
For our corpora, we will maintain references to Twitter and not to X throughout the manuscript, as Twitter was the name of the social network while the posts compilation was performed.}
retrieved from the filtered stream endpoint of the Twitter API. The use of tweet datasets as quantitative linguistic corpora has garnered significant interest for exploring lexical variation and entropy measurement applications in language studies. Twitter provided real-time, informal language data that reflects contemporary linguistic trends and variations~\cite{EisensteinDiffusionLexical2014,gonccalves2014, gonccalves2018,AbitbolSocioeconomicDependencies2018,GrieveMappingLexical2019,AlshaabiStorywranglerMassive2021, louf2021, tellez2022, louf2023, dunn2023syntactic},
useful for research in computational sociolinguistics. Recent studies have applied entropy measures to quantify the unpredictability and information content within tweet corpora, allowing for user classification~\cite{ghosh2011}, tweet veracity determination~\cite{paryani2017}, and sentiment analysis for stock market predictions~\cite{Kanavos2020}. 

Tweets are automatically classified by language using Google CLD2~\cite{cld2_github} with a threshold of 95\%. Consequently, any tweet that contains less than 95\% of its content in the language under study is discarded. 

The fourth corpus contains $2.27 \times 10^8$ tweets in English (TwEN) from 2022, geolocalized in the USA, Canada, Greenland, Bermuda, Saint Pierre and Michelon, and Great Britain. After preprocessing, this corpus consists of 1,800,000,005 tokens and 2,839,568 types, with 1,427,618 hapax legomena, which constitute 50\% of the total vocabulary.

Then, the fifth corpus (TwES) is a compilation of $3.32 \times 10^8$ tweets in Spanish from 2020 to 2022, geolocalized in Spanish-speaking countries (Spain and Latin America). After preprocessing, this dataset consists of 1,800,000,011 tokens and 4,305,331 types, of which 2,544,306 (59\%) are hapax legomena.

Finally, the sixth corpus comprises $3.41 \times 10^8$ tweets in Turkish (TwTR) from 2015 to 2022 geolocalized in the Republic of Türkiye. After preprocessing, the TWTR corpus consists of 1,800,000,004 tokens and 13,262,464 types, of which 8,157,836 (62\%) constitute hapax legomena.

We remark that the time periods of the three Twitter corpora are selected such that the token number is almost the same, which is good for comparison purposes. Further,
the increasing number of hapax legomena from English to Spanish and then to Turkish reflects their unique morphological typologies as explained above. Whereas Turkish formation of words is more flexible, the isolating characteristics of English makes word formation more rigid, the fusional mechanisms of Spanish lying approximately in the middle. This fact is reflected in the $TTR$ values in Table~\ref{tab:corpora}, where the flexibility of Turkish morphology increases the relative vocabulary available in comparison to Spanish or English, and also in the results of Section~\ref{sec:entropy}.

\begin{figure}[t]
  \centering
  \includegraphics[width=0.95\linewidth]{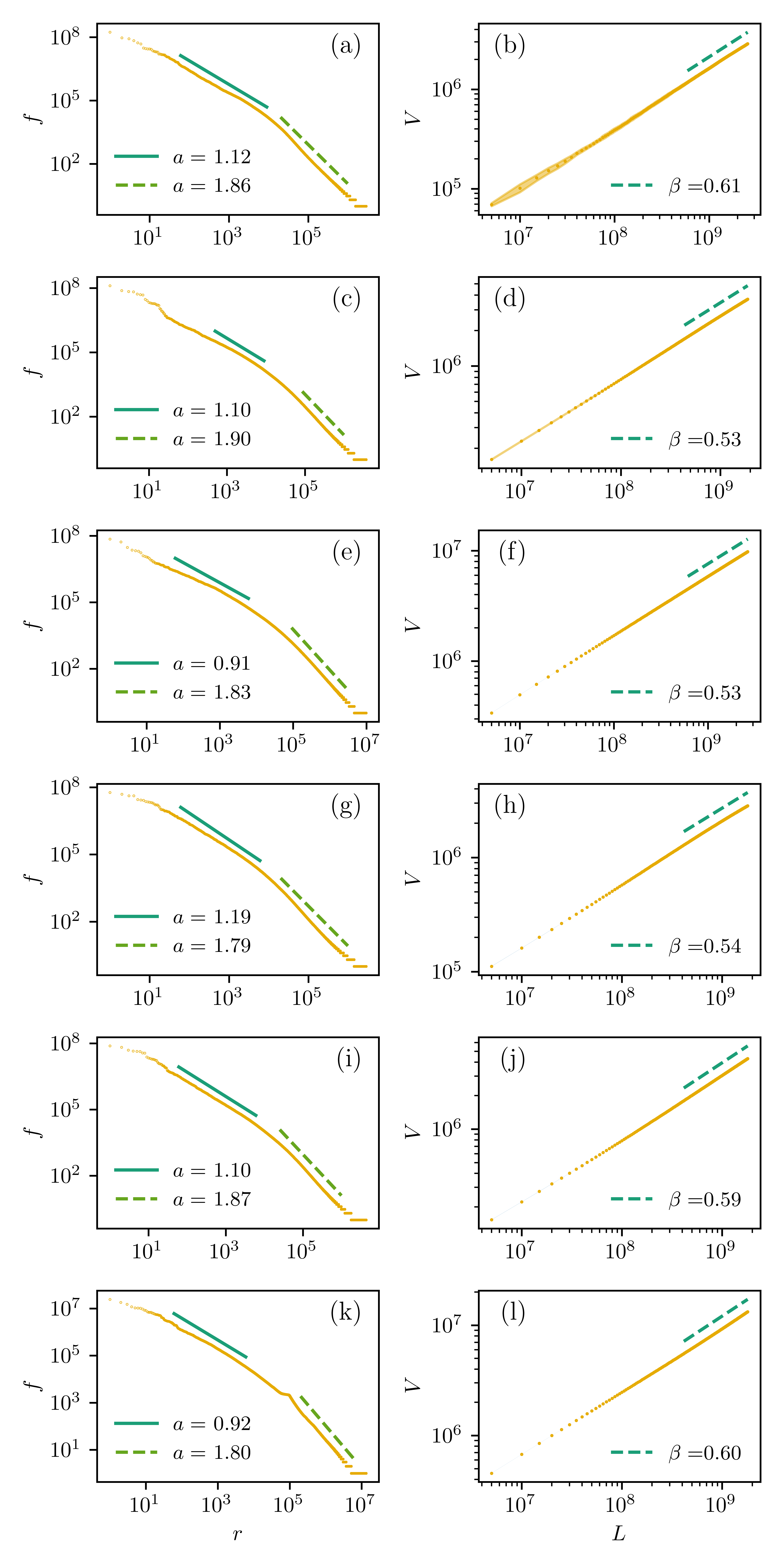}
  \caption{Frequency $f$ as a function of the rank $r$ (left) and vocabulary or lexicon $V$ as a function of text length $L$ (right) laws for the SPGC [(a), (b)], SPA [(c), (d)], TRCC100 [(e), (f)], TwEN [(g), (h)], TwES [(i), (j)], and TwTR [(k), (l)] corpora. The errors of the estimation of both the Zipf exponent $a$ and the Heaps exponent $\beta$ are lower than 0.001.}
  \label{fig:zhl}
\end{figure}

We begin by analyzing Zipf and Heaps laws, which are renowned for being fulfilled in large texts of natural language and have been shown to be related under certain assumptions~\cite{vanleijenhorst2005, Serrano2009_1, Lu2010, Eliazar2012, Fontclos2013, loreto2016dynamics}. We  examine Zipf and Heaps laws throughout the six corpora, as shown in Fig.~\ref{fig:zhl}. Zipf distribution for the frequency $f$ of a word $w$ with rank $r$ reads
\begin{equation}
\label{eq:zipf}
    f = \frac{k}{r^a},
\end{equation}
where $k$ is a normalization constant and $a=1$, although the empirically found exponent differs from $a=1$~\cite{Stanisz2024}. Here, the rank $r$ of a word is defined as its position in the list of all words when they are ordered by decreasing frequency. In other words, the most frequent word is assigned a rank of 1, the second most frequent a rank of 2, and so on. Approximations to Zipf law are observed throughout the six corpora, noting the existence of two different exponents~\cite{petersen2012,yamamoto2021negative}. When $r$ surpasses a critical value around $10^4-10^5$, the frequency of usage of words decreases faster than before. The first exponent is close to a perfect Zipf law ($a=1$), and corresponds to the rank range of the kernel vocabulary as described in the literature~\cite{Ferrer2001}. The second exponent would account for rare words, like neologisms, technical terms, variants of words pertaining to the kernel vocabulary, typos, etc.

On the other hand, Heaps (or Herdan) law~\cite{herdan1960,Heaps1978} relates the vocabulary or lexicon size (number of types) $V$ of a text to its length (number of tokens) $L$ as
\begin{equation}
\label{eq:heaps}
    V = \alpha L^\beta,
\end{equation}
where $\beta\in(0,1)$. This law dictates that the vocabulary size grows sublinearly as more text is analyzed within the corpus. As shown in Fig.~\ref{fig:zhl}, our six corpora fulfill
Heaps law remarkably well, over several decades of $L$.
The empirically found exponent is in the range between $0.53$ and $0.61$,
in agreement with previously reported observations
in corpora of similar sizes~\cite{bochkarev2014deviations}. 

\begin{table}[t] 
\caption{Approximate values for the number of tokens (or text length), $L$, the number of types (or vocabulary), $V$, and the type-to-token ratio, $TTR~=~V/L$, of the different corpora used in the analysis. We also show their language and source.\label{tab:corpora}}
\begin{tabular}{cccccc}
\hline \hline
Corpus & $L~ \left(10^9\right)$ & $V~ \left(10^6\right)$ & $TTR \left(10^{-3}\right)$ & Lang. & Src. \\
\hline
SPGC    & $2.62$ & $3.1$ & $1.2$ & English  & Books    \\
SPA     & $1.89$ & $3.7$ & $2.0$ & Spanish  & Websites \\
TRCC100 & $2.64$ & $9.8$ & $3.7$ & Turkish  & Websites \\
TwEN    & $1.80$ & $2.8$ & $1.6$ & English  & Twitter  \\
TwES    & $1.80$ & $4.3$ & $2.4$ & Spanish  & Twitter  \\
TwTR    & $1.80$ & $13.3$ & $7.4$ & Turkish  & Twitter  \\
\hline \hline
\end{tabular}
\end{table}

\section{Entropy}
\label{sec:entropy}

In this section, we present the word entropy results for the corpora, showing how well it aligns with an analytical expression based on Zipf and Heaps laws in the limit of a very large text length. Let us first examine the empirical entropy data.

\subsection{Entropy measurement}

The Shannon entropy in Eq.~\eqref{eq:shannonh} may be estimated using different methods. The simplest one is the so-called plug-in (PI) estimator, which consists of assigning the normalized empirical frequency to the exact probability to compute the Shannon entropy, i.e., it consists of plugging the normalized empirical frequency into the sum in Eq.~\eqref{eq:shannonh}.  

\begin{figure}[t]
  \centering
  \includegraphics[width=0.95\linewidth]{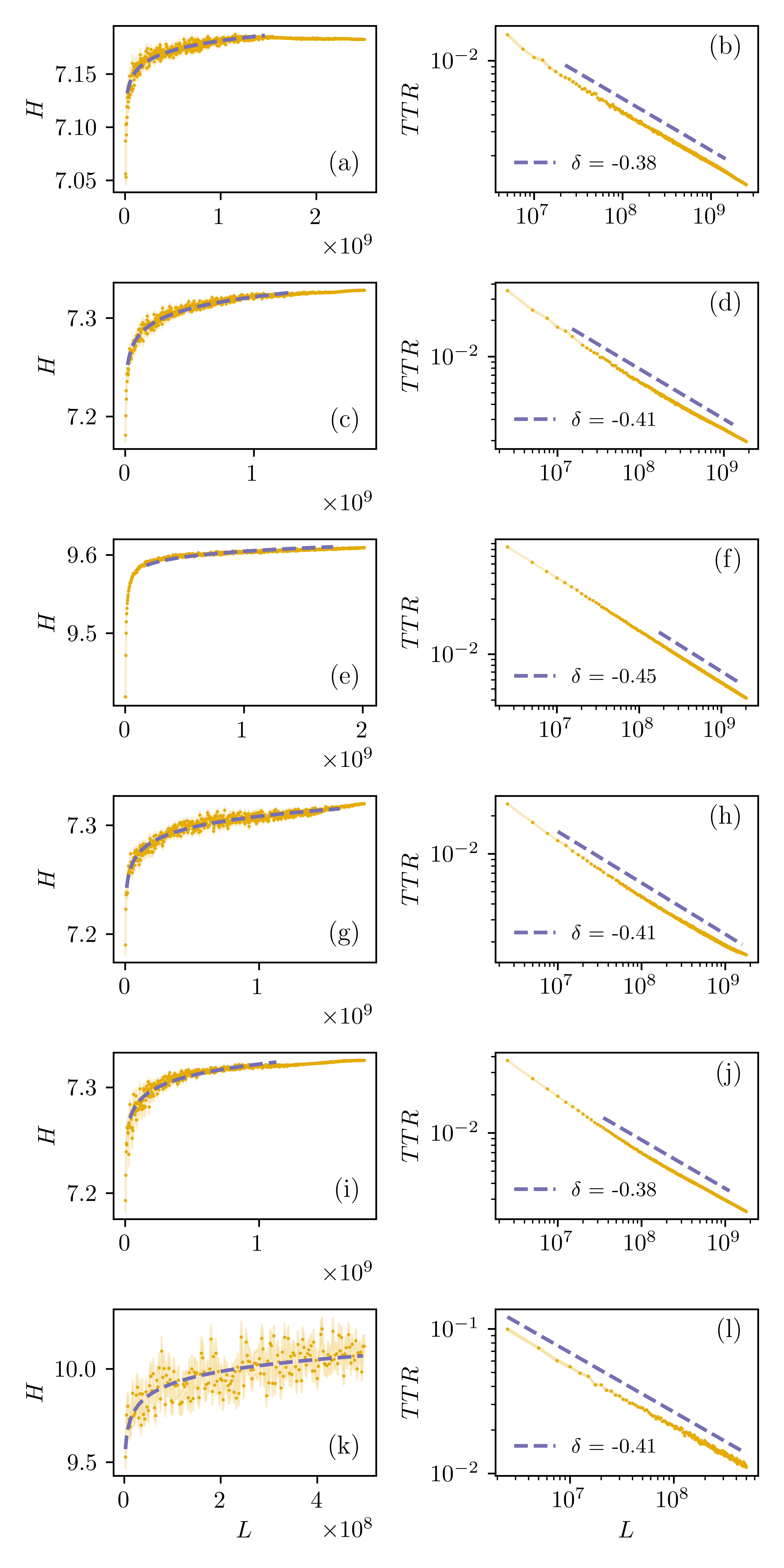}
  \caption{Measured entropy $H$ (left) and $TTR$ in terms of text length $L$ (right) for the SPGC [(a), (b)], SPA [(c), (d)], TRCC100 [(e), (f)], TwEN [(g), (h)], TwES [(i), (j)], and TwTR [(k), (l)] corpora. For $H$ and $TTR$ we plot the mean value and its error. In dashed lines, we fit the data to the corresponding analytical expressions in Eq.~\eqref{eq:hfitl} and Eq.~\eqref{eq:ttr_l}. The error in the estimation of $\delta$ is always lower than 0.001.}
  \label{fig:h_ttr}
\end{figure}

In Fig.~\ref{fig:h_ttr}(left), we present entropy measurements as a function of $L$ for the six gigaword corpora using the PI estimator. The detailed measurement procedure is explained in Appendix~\ref{app:computation}. In the case of the TwTR corpus depicted in Fig.~\ref{fig:h_ttr}(k), we restrict our analysis to $L \leq 5\times 10^8$ due to a lack of reliable statistics beyond that point. Our observations  reveal a clear universal dependence of the entropy on $L$ across all corpora and registers. We will later dive on the details of this functional form.

Interestingly, the data also shows slight quantitative differences related to both language and register. We recall that our corpora include three languages (English, Spanish, and Turkish) and three main sources: books (formal register), online media (combination of both formal and informal registers), and microblogging posts (mostly informal register).

Analyzing by language, we observe that Turkish [Fig.~\ref{fig:h_ttr}(e) and Fig.~\ref{fig:h_ttr}(k)] consistently shows the highest entropy, followed by Spanish [Fig.~\ref{fig:h_ttr}(c) and Fig.~\ref{fig:h_ttr}(i)], and then English [Fig.~\ref{fig:h_ttr}(a) and Fig.~\ref{fig:h_ttr}(g)]. This hierarchy difference in entropy values is explained by the morphological structure of each language. Turkish, with its complex, agglutinative morphology, leads to a high number of unique word forms. Consequently, the probability distribution of words in Turkish is more dispersed, leading to higher entropy. By comparison, Spanish morphology is less agglutinative and more isolating than Turkish, while English is the most isolating of the three. This progression reflects the observed differences in entropy among the languages, which is in accordance to previous descriptions in the literature~\cite{Bentz2017}.

Analyzing by register, we find that within any given language, entropy increases with register informality. Informal registers, such as microblogging posts [in our case, tweets in Fig.~\ref{fig:h_ttr}(g), Fig.~\ref{fig:h_ttr}(i) and Fig.~\ref{fig:h_ttr}(k)], typically exhibit higher entropy than more formal registers like books [Fig.~\ref{fig:h_ttr}(a)] or online media [Fig.~\ref{fig:h_ttr}(c) and Fig.~\ref{fig:h_ttr}(e)]. This pattern arises because informal writing tends to be more spontaneous and less constrained by standardized grammar or vocabulary. In informal contexts, language users often introduce unique expressions, slang, abbreviations, and varied syntactic structures, resulting in a broader range of distinct word forms. This variability expands the vocabulary probability distribution, thus increasing entropy. In contrast, formal registers tend to adhere to a more standardized language use, with recurring vocabulary and grammar structures that limit entropy growth.

\subsection{Asymptotic expression}

To gain further insight into the the entropy dependence on $L$, we now examine Eq.~\eqref{eq:shannonh} in the large vocabulary limit since we are dealing with massive corpora.

The Shannon entropy of the Zipf distribution given by Eq.~\eqref{eq:zipf} is obtained straightforwardly and reads
\begin{equation}
\label{eq:hzipf}
    H = \frac{a}{K_{V,a}} \sum\limits_{r=1}^{V} \frac{\ln r}{r^a} + \ln K_{V,a},
\end{equation}
where 
\begin{equation}
    \notag
    K_{V,a} = \sum\limits_{n=1}^{V} n^{-a}.
\end{equation}

For large vocabularies, in virtue of the Euler-Maclaurin formula~\cite{Apostol1999} we have that
\begin{equation}
    K_{V,1} \sim \ln V,
\end{equation}
and as
\begin{equation}
    \sum\limits_{r=1}^{V} \frac{\ln r}{r}  =  \gamma_1 - \gamma_1 (V + 1),
\end{equation}
where $\gamma_1$ refers to the first Stieltjes constant and $\gamma_1(x)$ to the first generalized Stieltjes constant~\cite{Liang1972}, we end up with
\begin{equation}
    \sum\limits_{r=1}^{V} \frac{\ln r}{r} \sim  \frac{\left(\ln V\right)^2}{2}
\end{equation}
for large $V$~(see leading order of Eq.~(1.12) in Ref.~\cite{Coffey2014} or Eq.~(3.1) of~\cite{Liang1972}). Compiling the above relations, one arrives at~\cite{jones1979, Corominas2010, Visser2013}
\begin{equation}
\label{eq:hfit}
    H \sim \frac{1}{2}\ln V + \ln \ln V.
\end{equation}

Throughout the manuscript, we use the symbol $\sim$ to denote that two functions are asymptotically equivalent. Specifically, if  
\begin{equation}
\lim_{n \to \infty} \frac{f(n)}{g(n)} = 1,
\end{equation}  
we write $f \sim g$ as $n \to \infty$.

Equation~\eqref{eq:hfit} is an asymptotic result based on a Zipf exponent~$a = 1$, but, as previously mentioned, our empirical results in Fig.~\ref{fig:zhl} indicate the existence of two regimes with two different exponents. Whereas the first exponent can be safely approximated by~1, the second exponent is larger than~1. However, we now show that the regime associated with the second exponent has a negligible contribution to the entropy values in relation to the regime associated with the first exponent.

\begin{figure}[t]
  \centering
  \includegraphics[width=0.99\linewidth]{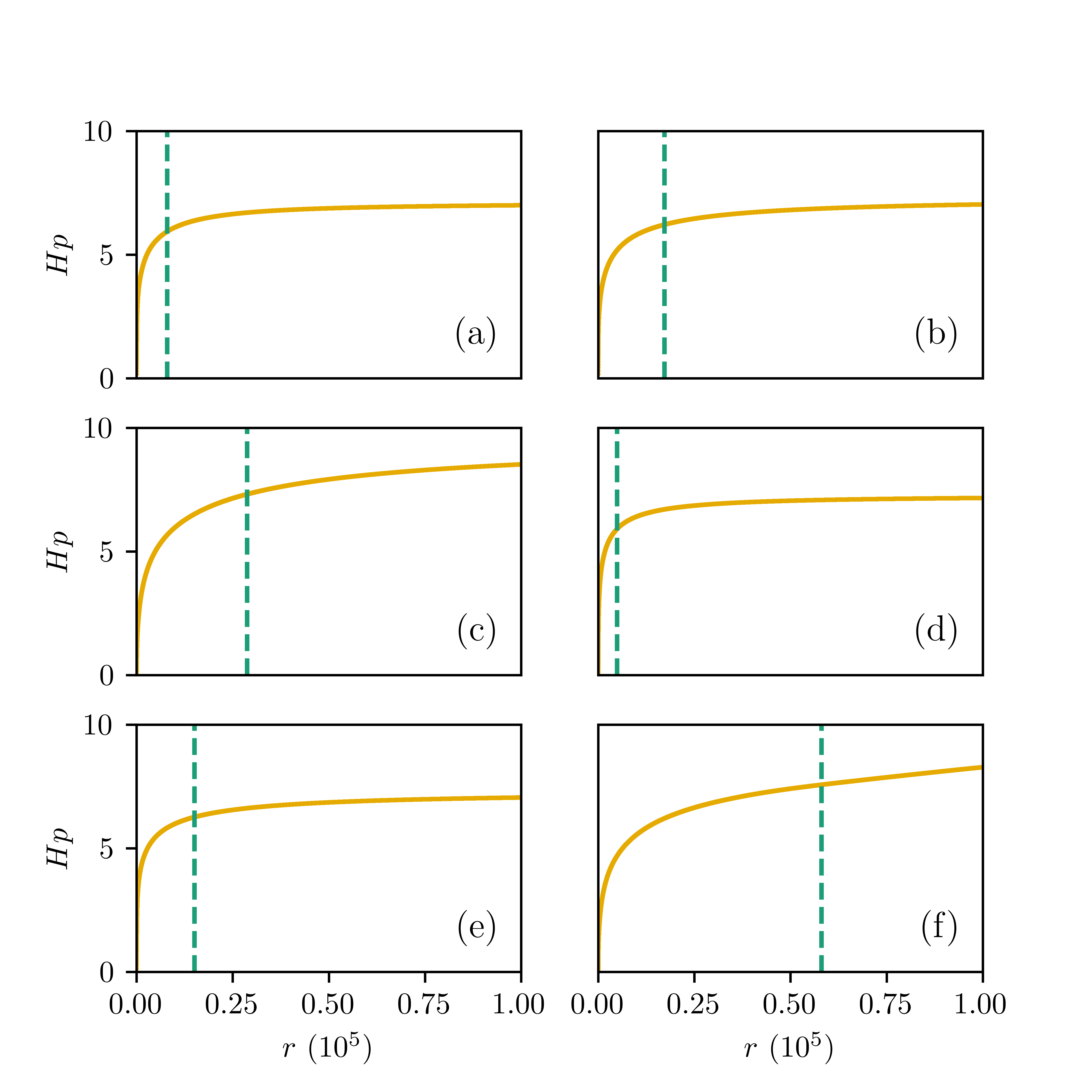}
  \caption{Incremental partial entropy, $H_p$, in terms of the ordered ranks, as in Eq.~\eqref{eq:incr_entropy}, for the SPGC (a), SPA (b), TRCC100 (c), TwEN (d), TwES (e), and TwTR (f) corpora. We show as dashed vertical lines the position of the rank in which the Zipf law changes exponent.}
  \label{fig:partial_h}
\end{figure}

To see this, let us consider the incremental partial entropy, 
\begin{equation}
    \label{eq:incr_entropy}
    H_p (r) = -\sum\limits_{r'=1}^r p_{r'} \ln p_{r'},
\end{equation}
where $p_{r'}$ corresponds to the empirical probability of the word with rank $r'$. We plot $H_p(r)$ in Fig.~\ref{fig:partial_h}. We also indicate with a vertical dashed line the rank in which the Zipf law changes exponent. Clearly, the ranks that can be fit with $a\simeq 1$ account for around $80\%$ of the observed entropy values or more, as shown in  Appendix~\ref{app:2zipfcont} using different proxies. For this reason, the $a = 1$ assumption is a good approximation.

In Eq.~\eqref{eq:hfit} we express entropy in terms of vocabulary,~$V$. However, as shown in Appendix~\ref{app:computation}, our control variable during the entropy measurement is text length,~$L$. Thus, we express Eq.~\eqref{eq:hfit} in terms of $L$ for large samples using Heaps law in Eq.~\eqref{eq:heaps} and find
\begin{equation}
\label{eq:hfitl}
    H \sim \frac{\beta}{2} \ln L + \ln\ln L,
\end{equation}
where the dependence on $\alpha$ is negligible in the limit of large sample sizes. Even though the first term of Eq.~\eqref{eq:hfitl} dominates over the second term for $L \gg 1$, it is relevant to note that our datasets involve text lengths on the order of $10^9$ tokens and therefore the contribution of the second term can not be neglected.

\subsection{Validation}

We now proceed to compare Eq.~\eqref{eq:hfitl} to the entropy measurements from the six corpora. This fitting process will allow us to assess how well the predicted behavior aligns with observed entropy values across different language-register combinations.

Given Eq.~\eqref{eq:hfitl}, we fit the data to
\begin{equation}
\label{eq:fith}
    H_\text{fit}\left( L \right) = p_1 \left( \frac{\beta}{2} \ln L + \ln\ln L \right) + p_2
\end{equation}
where $p_1$ and $p_2$ are fitting parameters, independent of $L$, for the six datasets, in order to align the asymptotic approximation to the empirical data. In other words, Eq.~\eqref{eq:fith} is a semi-empirical approach that retains the same functional behaviour of $H(L)$ for all corpora with $p_1$ and $p_2$ being specific for each dataset. This precisely follows the spirit of other statistical laws~\cite{Altmann2016}. Also, the $\beta$ values are the ones obtained in Fig.~\ref{fig:zhl}.

The results of a nonlinear least squares (NLS) fitting~\cite{Vugrin2007} are shown with dashed lines in Fig.~\ref{fig:h_ttr} and numerically in Table~\ref{tab:fithl}, where we also present a few metrics to evaluate the goodness of the fit for Eq.~\eqref{eq:fith} to the data, namely, the coefficient of determination, $\rho^2$, the Spearman correlation coefficient, $\rho_\mathrm{s}$, the distance correlation coefficient $\rho_\mathrm{d}$ (a measure suitable for nonlinear dependences~\cite{Gabor2007,satra2014}), and the $p$-value. We see a clear agreement between data and the fit of Eq.~\eqref{eq:fith}, except for the TWTR corpus due to its poor statistics, as mentioned above. Specifically, the worst values of the metrics in Table~\ref{tab:fithl} (without considering the TwTR corpus) are $\rho^2_\mathrm{min} = 0.91$, $r_\mathrm{s, min} = 0.93$, and~$\rho_\mathrm{d, min} = 0.90$.

\begin{figure}[t]
  \centering
  \includegraphics[width=0.9\linewidth]{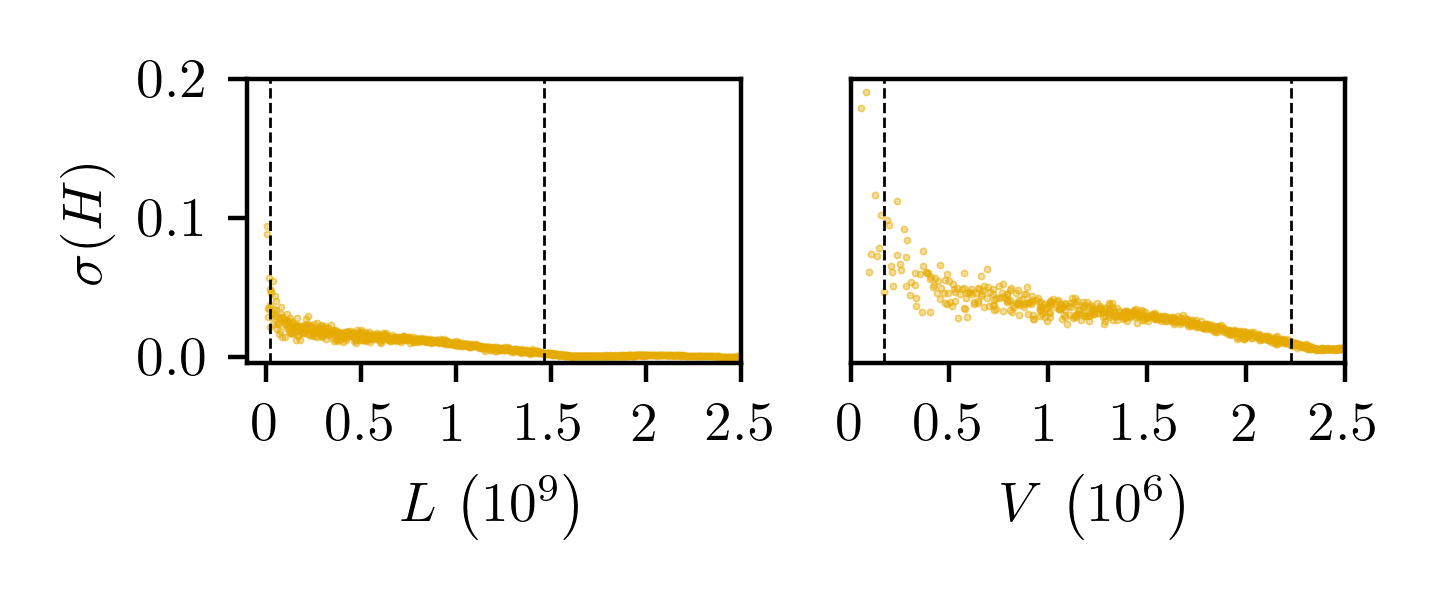}
  \caption{Range selection for the fits of the SPGC in Fig.~\ref{fig:h_ttr} studying the standard deviation for the entropy distributions $\sigma(H)$ in terms of vocabulary $V$ or text length $L$. In vertical dashed lines we show the interval $0.0025 < \sigma_H (V) < 0.025$. We choose this interval to avoid the regions in which the standard deviation is either too high due to the noise caused by the low size of the samples, or too low due to the sample size getting close to the corpus size.}
  \label{fig:range_selection}
\end{figure}

Our selection for the range of $L$ on which we perform the fit is based on an analysis of $\sigma(H)$, the standard deviation of $H$.
An example is shown in Fig.~\ref{fig:range_selection}. We observe that $\sigma(H)$ is large for small $L$ due to noisy statistics. On the other hand, $\sigma(H)$ is too small for large $L$, where the sample size approaches the corpus size and the statistics are then not reliable. Therefore, we take $0.0025~<~\sigma_H~<~0.025$ as a reasonable interval for the fit. The rest of the cases can be found in Appendix~\ref{app:range}. 

\begin{table}[t] 
\caption{Results of the fit of Eq.~\eqref{eq:fith} to the corpora shown in Fig.~\ref{fig:h_ttr}. As evaluations of the goodness of the fit, we show the coefficient of determination, $\rho^2$, the Spearman correlation coefficient, $\rho_\mathrm{s}$, and the distance correlation, $\rho_\mathrm{d}$. In all cases, $p\text{-value}<0.0001$.\label{tab:fithl}}
\begin{tabular}{cccccc}
\hline \hline
Corpus & $p_1$ & $p_2$ & $\rho^2$ & $\rho_\mathrm{s}$ & $\rho_\mathrm{d}$ \\
\hline
SPGC    & 0.04  & 6.83  & 0.91  & 0.94  & 0.90  \\
SPA     & 0.06  & 6.85  & 0.97  & 0.98  & 0.93 \\
TRCC100 & 0.03  & 9.32  & 0.92  & 0.98  & 0.90  \\
TwEN    & 0.05  & 6.91  & 0.93  & 0.93  & 0.90  \\
TwES    & 0.05  & 6.91  & 0.91  & 0.97  & 0.90  \\
TwTR    & 0.27  & 7.69  & 0.58  & 0.74  & 0.72   \\
\hline \hline
\end{tabular}
\end{table}

We recall that Figs.~\ref{fig:h_ttr}, \ref{fig:partial_h} and~\ref{fig:range_selection}
were obtained using the PI estimator for the entropy estimation. However, there exist other estimators~\cite{Bentz2017}, such as the Nemenman–Shafee–Bialek (NSB) estimator~\cite{nemenman2001entropy}, which has been shown to be less biased than the plug-in estimator and closer to real entropy~\cite{arora2022,juani}. In Appendix~\ref{app:nsb} we recalculate the entropy
and its deviation with the NSB estimator and find no significant deviations.


\section{Type-token ratio}
\label{sec:ttr}

With the data of $L$ and $V$ we can compute the type-token ratio as defined in Eq.~\eqref{eq:ttr}. The results are plotted in Fig.~\ref{fig:h_ttr}(right), showing a clear decreasing dependence of $TTR$ on $L$.

This dependence can be understood from the sublinear relation between vocabulary and text length given by Heaps law, as observed in Fig.~\ref{fig:zhl}. Using Eq.~\eqref{eq:heaps}, we can express $TTR$ in terms of either the text length or the vocabulary, which results in a power law. Thus,
\begin{equation}
\label{eq:ttr_l}
    TTR = \alpha^{\frac{1}{\beta}} V^\frac{\beta-1}{\beta} = \alpha L^{\beta - 1} \sim L^\delta,
\end{equation}
which is valid asymptotically for $L\gg1$.

Equation~\eqref{eq:ttr_l} is in complete agreement with the data as shown in Fig.~\ref{fig:h_ttr}(right) and in Table~\ref{tab:fitttrl}. The low values for~$|\delta~-~\left( \beta-1 \right)|$ in Table~\ref{tab:fitttrl} support the validity of Heaps law to understand the TTR dependence on text length in the limit of large $L$. 

It is worth noting that there are multiple methods for calculating the type-token ratio. Throughout this work, we employ an over-all approach, which is explicitly influenced by text length. Alternative metrics, such as mean-segmental approaches, are designed to avoid scaling with sample size
. It is important to note that if we had used mean-segmental methods, our results would have been significantly different. This point is discussed further in Section~\ref{sec:conclusions}.

\begin{table}[t] 
\caption{Results for the exponent $\delta$ of the power law relation between the type-token ratio and the text length. The error in the estimation of $\delta$ is always lower than 0.001. For assessing the goodness of the fit, we provide the coefficient of determination, $\rho^2$, the Spearman correlation coefficient, $\rho_\mathrm{s}$, and the distance correlation, $\rho_\mathrm{d}$. In all cases, $p\text{-value}<0.0001$. To check its accordance with Heaps law in Eq.~\eqref{eq:heaps}, we show the absolute difference of the exponents,~$|\delta - \left( \beta-1 \right)|$. We show the value of $\beta$ in Fig.~\ref{fig:zhl}.\label{tab:fitttrl}}
\begin{tabular}{cccccc}
\hline \hline
Corpus & $\delta$ & $\rho^2$ & $\rho_\mathrm{s}$ & $\rho_\mathrm{d}$ & $|\delta-\left(\beta - 1\right)|$ \\
\hline
SPGC    & -0.38 & 1.00 & -1.00 & 0.92 &  0.01 \\
SPA     & -0.41 & 0.98 & -1.00 & 0.90 &  0.06 \\
TRCC100 & -0.45 & 1.00 & -1.00 & 0.96 &  0.02 \\
TwEN    & -0.41 & 1.00 & -1.00 & 0.89 &  0.05 \\
TwES    & -0.38 & 1.00 & -1.00 & 0.93 &  0.02 \\
TwTR    & -0.41 & 1.00 & -0.96 & 0.86 &  0.01 \\
\hline \hline
\end{tabular}
\end{table}

\section{Entropy and type-token ratio}
\label{sec:entropyandttr}

\begin{figure}[b]
  \centering
  \includegraphics[width=0.7\linewidth]{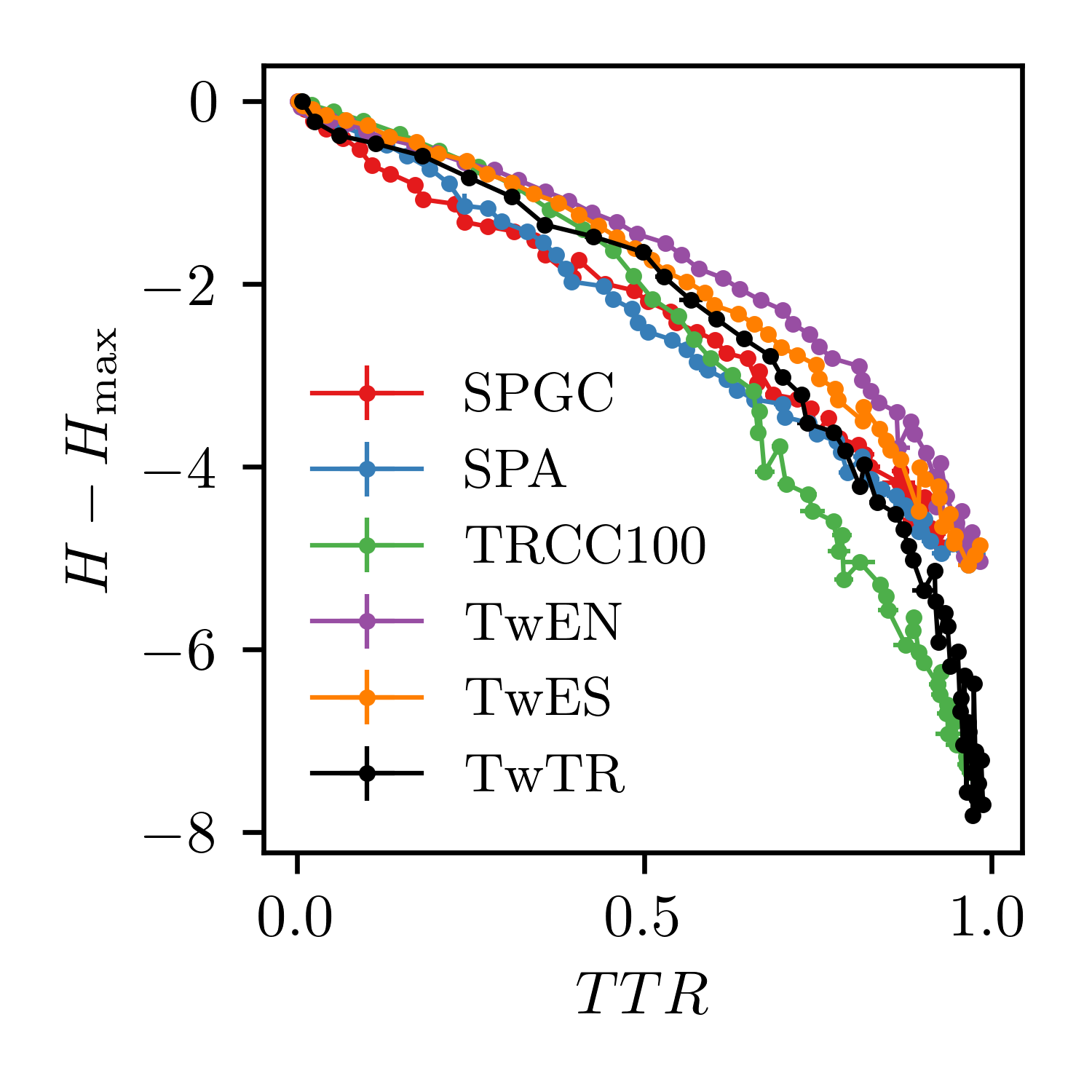}
  \caption{Deviations from maximum entropy, $H_\mathrm{max}$, in terms of $TTR$. Data is shown for the SPGC (red), SPA (blue), TRCC100 (green), TwEN (purple), TwES (orange), and TwTR (black) corpora. Each point represents the mean over 25 random samples of a given length $L$. The error is represented but is barely seen due the scale of the entropy values.
  }
  \label{fig:h_ttr_huge}
\end{figure}

We now compare the two distinct metrics for lexical diversity discussed in~Secs.~\ref{sec:entropy} and~\ref{sec:ttr}. In Fig.~\ref{fig:h_ttr_huge}, we show the entropy and $TTR$ values for text portions of varying lengths in the studied corpora, where each point corresponds to a specific $L$ value. We present the entropy data as deviations from the maximum value, i.e., $H_\mathrm{max}$, which corresponds to the value of $H$ for the largest $L$, which is set to be $L=1.8 \times 10^9$ for all the corpora in Fig.~\ref{fig:h_ttr_huge} to allow for graphic comparison. From Fig.~\ref{fig:h_ttr}, we already know that $TTR$ decreases as $L$ increases, while entropy shows the opposite trend, suggesting an expected negative correlation between entropy and type-token ratio. The empirical functional relationship between entropy and type-token ratio that we find is shown in Fig.~\ref{fig:h_ttr_huge}. The alignment observed in $H-H_\mathrm{max}$ among the corpora suggests a similar behavior, especially near the asymptotic limit $TTR \rightarrow 0$. This can be understood by deriving an expression for $H(TTR)$ for large vocabularies. By substituting Eq.~\eqref{eq:ttr_l} in Eq.~\eqref{eq:hfitl} and noting that $0<\beta<1$, we arrive to
\begin{align}
    \label{eq:h_ttr_nofit}
    H \sim \frac{\beta}{2\left(1-\beta\right)}\ln TTR^{-1} + \ln\ln TTR^{-1},
\end{align}
which is valid for $TTR \ll 1$. 

\begin{figure}[t]
  \centering
  \includegraphics[width=0.95\linewidth]{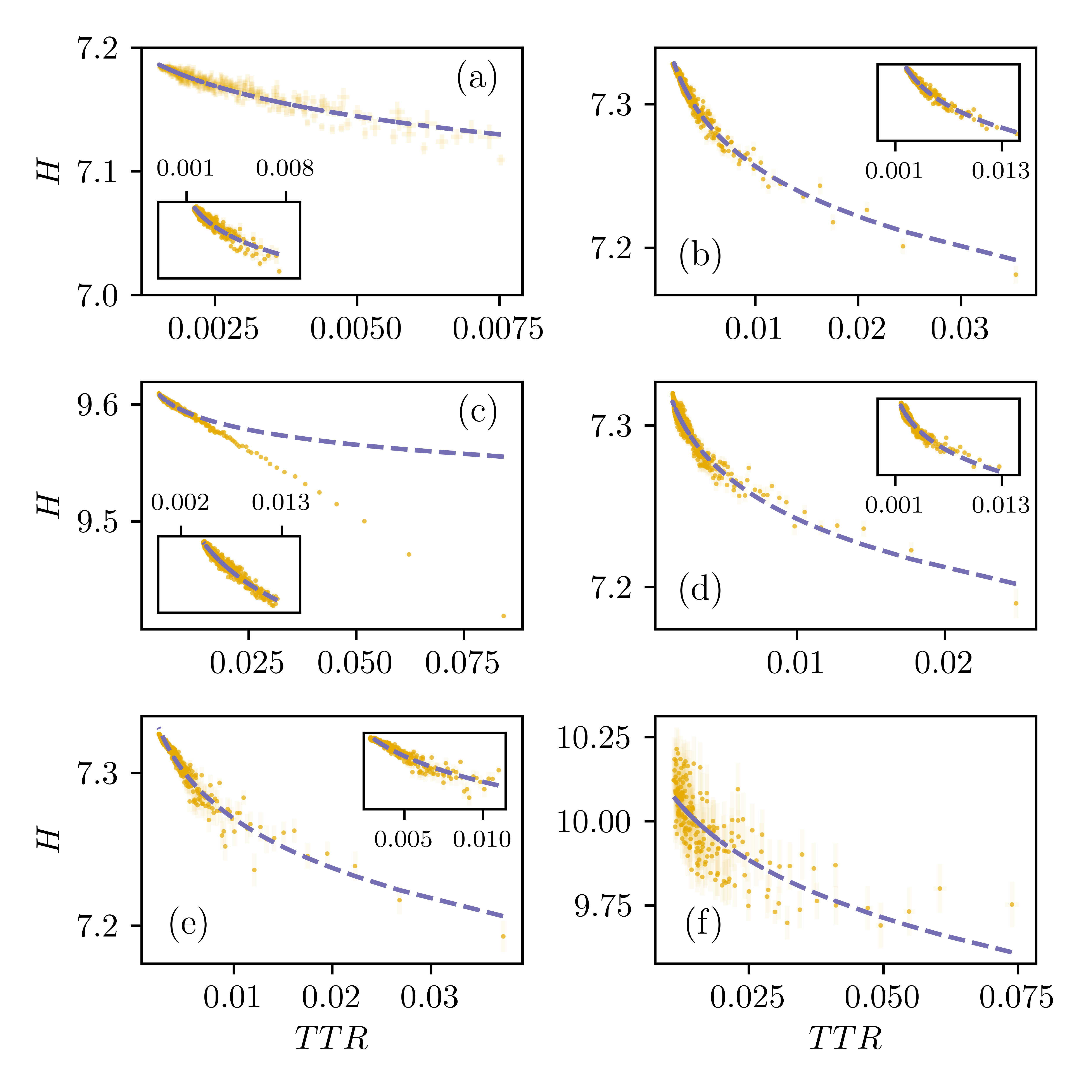}
  \caption{$H$ in terms of $TTR$ for the SPGC (a), SPA (b), TRCC100 (c), TwEN (d), TwES (e), and TwTR (f) corpora for $L > 2.5\cdot 10^6$ tokens. With a dashed line we plot the fit to Eq.~\eqref{eq:h_ttr}, whose goodness is assessed in Table~\ref{tab:fit_h_ttr}. Inside the insets we show the data and the fit in the confidence interval, as exemplified in Fig.~\ref{fig:range_selection}.}
  \label{fig:h_ttr_fit}
\end{figure}

In order to validate Eq.~\eqref{eq:h_ttr_nofit}, we fit
\begin{align}
    \label{eq:h_ttr}
    H_\text{fit}\left(TTR \right) &= p_{3} \left[\frac{\beta}{2\left(\beta-1\right)}\ln TTR^{-1} \right. \\
    \notag
    &\left. + \ln\ln TTR^{-1}\right] + p_{4},
\end{align}
to the data, $p_{3}$ and $p_{4}$ being fitting parameters. Using the values for $\beta$ extracted from the fits of Fig.~\ref{fig:zhl}, we obtain the NLS fits shown in Fig.~\ref{fig:h_ttr_fit} (dashed lines) and in Table~\ref{tab:fit_h_ttr} using the plug-in  estimator. These results show an overall good agreement between the asymptotic expression in Eq.~\eqref{eq:h_ttr} and the data. Equation~\eqref{eq:h_ttr} inherits the validity for large $L$ from Eq.~\eqref{eq:hfitl}, and hence we observe an asymptotic validity of the relation in Eq.~\eqref{eq:h_ttr_nofit}. This fit clarifies the relationship between $H$ and $TTR$ for small values of $TTR$, which is based on logarithmic functions. This relationship is expressed by Eq.~\eqref{eq:h_ttr_nofit} and depends on the validity of Zipf and Heaps laws. Similar results using the NSB estimator are shown in Appendix~\ref{app:nsb}.

\begin{table}[t] 
\caption{Evaluation of the goodness of the fit for Eq.~\eqref{eq:h_ttr}. We provide the values of the determination coefficient, $\rho^2$, the Spearman correlation coefficient, $\rho_\mathrm{s}$, and the distance correlation, $\rho_\mathrm{d}$. In all cases, $p\text{-value}<0.0001$.\label{tab:fit_h_ttr}}
\begin{tabular}{cccccc}
\hline \hline
Corpus & $p_3$ & $p_4$ & $\rho^2$ & $\rho_\mathrm{s}$ & $\rho_\mathrm{d}$ \\
\hline
SPGC    & 0.04 & 6.93 & 0.91  & -1.00 & 0.94 \\
SPA     & 0.61 & 5.61 & 0.97  & -1.00 & 0.97 \\
TRCC100 & 0.24 & 8.94 & 0.96  & -1.00 & 1.00 \\
TwEN    & 0.49 & 5.95 & 0.91  & -1.00 & 0.98 \\
TwES    & 0.43 & 6.13 & 0.92  & -1.00 & 0.99 \\
TwTR    & 2.48 & 3.03 & 0.60  & -0.99 & 0.98 \\
\hline \hline
\end{tabular}
\end{table}

\section{Conclusions}
\label{sec:conclusions}
In this work, we conducted an empirical analysis of entropy and type-token ratio using data from six corpora in three languages: English, Spanish, and Turkish, each representing different morphological typologies. These corpora, sourced from books, media outlets, and Twitter (now X), contain over $10^9$ word tokens, enabling robust statistical analysis. Our key finding is a consistent functional relationship between $H$ and $TTR$ across all datasets, as seen in Fig.~\ref{fig:h_ttr_huge}.

To better understand this relationship, we derived Eq.~\eqref{eq:hfitl}, an analytical expression for $H$ in the limit of large text lengths based on Zipf law. We also know that the dependence of $TTR$ on text length follows Eq.~\eqref{eq:ttr_l}, in virtue of Heaps law. This leads to Eq.~\eqref{eq:h_ttr_nofit}, a function that directly links $H$ and $TTR$, aligning with our empirical observations. The asymptotic alignment between corpus data and our analytical expression reinforces a functional dependence between entropy and type-token ratio.

This is one of our main findings that has implications for precise statistical modeling of large-scale natural language datasets. Even though, in principle, word entropy and type-token ratio might be seen as independent metrics, they are linked for a text of a sufficiently-large given length due to statistical laws of natural language. The asymptotic relation in Eq.~\eqref{eq:h_ttr_nofit} links a macro-level variable in the complex system of language (type-token ratio) with a micro-level measure of information (entropy) which is influenced by deeper structural patterns within the text. 

A word of caution is here in order. Our results depend on the over-all approach used when computing the type-token ratio. Using instead a mean-segmental approach would require fixing a specific text length~\cite{Johnson1944}. This adjustment would alter the results presented in Section~\ref{sec:entropyandttr} and invalidate the use of the variable $L$ for analytically linking entropy and the type-token ratio. Then, since both metrics reflect lexical richness, a positive correlation would be found between the mean-segmental (or standardized) type-token ratio and entropy, as shown in the literature~\cite{Bentz2016}. In this study, we do not obtain this positive correlation because we use the over-all approach to compute the type-token ratio. 

Our research is not without its limitations. First, as mentioned in the corpora description in Section~\ref{sec:data}, we use a pragmatical definition of words. There are others, like the lexical definition, in which words represent a base form or root, disregarding inflectional changes like verb conjugations. This causes that the entropy we employed is higher than the one we would have obtained in a lemmatized analysis, as the lexical definition of word consolidates multiple forms of each word. This consolidation results in fewer unique items and a more predictable word distribution, i.e., a lower entropy. We have also not accounted for the existence of homographs. In this sense, our entropy measurements are systematically lower than the ones that we would have obtained considering homographs as separate words, as the latter increase the available vocabulary, and hence the diversity and unpredictability. Further interesting research could study the differences in entropy between languages including lemmatized corpora and considering homographs.

Second, our analysis is limited to unigram entropy, which only considers the probability of individual words without accounting for dependencies between them. Written texts, however, often exhibit correlations between words~\cite{ebeling1994entropy,Altmann2012corr,montemurro2002long,drozdz2016quantifying,tanaka2016long,sanchez2023ordinal}. Such correlations can influence the overall structure and predictability of the text, reducing entropy when sequences are more predictable or increasing it when language patterns are more varied. Therefore, while unigram entropy provides valuable insights into vocabulary diversity, it does not capture the complexity introduced by these word-to-word correlations. To fully understand text entropy, future analyses should consider n-gram or sequence-based entropy measures, defining an $n$-gram $TTR$, which incorporate these dependencies and better reflect the inherent structure of written language.

Importantly, we did not include punctuation marks in our analysis to stay consistent with the text-filtering approach used in \cite{Gerlach2020}. However, this is an interesting direction for future work. Notably, Ref.~\cite{Kulig2017} shows that punctuation marks behave statistically like common words and can even improve the fit to Zipfian distributions.

Also, the analysis of Turkish Twitter data, from the TwTR corpus, presented specific challenges. Along with increased data dispersion, we lacked reliable statistics for cases where $ L > 5 \times 10^8 $, resulting in somewhat less accurate fits compared to those for other corpora and languages, though still reasonably robust. Future work that includes more comprehensive Twitter statistics for Turkish could enhance these findings.

We in addition remark that Eq.~\eqref{eq:h_ttr_nofit} is asymptotic, i.e., only valid for small $TTR$. Further studies should try to derive analytical expressions for $H(TTR)$ that shed light in the regimes corresponding to both intermediate and large $TTR$ values.

Finally, it would be interesting to investigate in future work whether the functional relationship found here between two diversity measures also holds in datasets from other fields, such as ecology and genetics, where Zipfs and Heaps laws have been observed~\cite{Camacho2001, sole2006}. Since our derivation shows that the entropy–TTR connection follows from these two laws, one would expect the same relationship to hold in any domain where they apply. However, our analysis has focused exclusively on language data, and testing this in other contexts could help assess the robustness and limitations of the connection.

\begin{acknowledgments}
We thank M. A. Tugores for insightful discussions about the corpora analysis, and J. De Gregorio and E. G. Altmann for valuable comments and references. This work was partially supported by the Spanish State Research Agency
(MICIU/AEI/10.13039/501100011033) and FEDER (UE) under project APASOS (PID2021-122256NB-C21) and the Mar{\'\i}a de Maeztu project CEX2021-001164-M.
\end{acknowledgments}

\appendix 

\section{Computation of the entropy and TTR}
\label{app:computation}

For each corpus described in Section~\ref{sec:data}, we compile the totality of texts and create a single document with all the content of the corpus. For the analysis of the six corpora, we compute entropy, $H$, vocabulary, $V$, and type-token ratio, $TTR$, of several partitions of different lengths of the compiled corpora of total length  $L_\text{T}$.

Given a compiled corpus of length $L_\text{T}$, we consider fragments with lengths $l \in [2.5\times 10^6,L_\text{T}]$ in steps of $\Delta l = 2.5 \times 10^6$ tokens. For each value of $l$, we take 25 random fragments of $l$ tokens from the total of $L_\text{T}$ tokens and compute the distributions of $H$, $V$, and $TTR$. The codes used for the analysis of the corpora are available in Ref.~\cite{repo}.

\begin{figure}[t]
  \centering
  \includegraphics[width=0.95\linewidth]{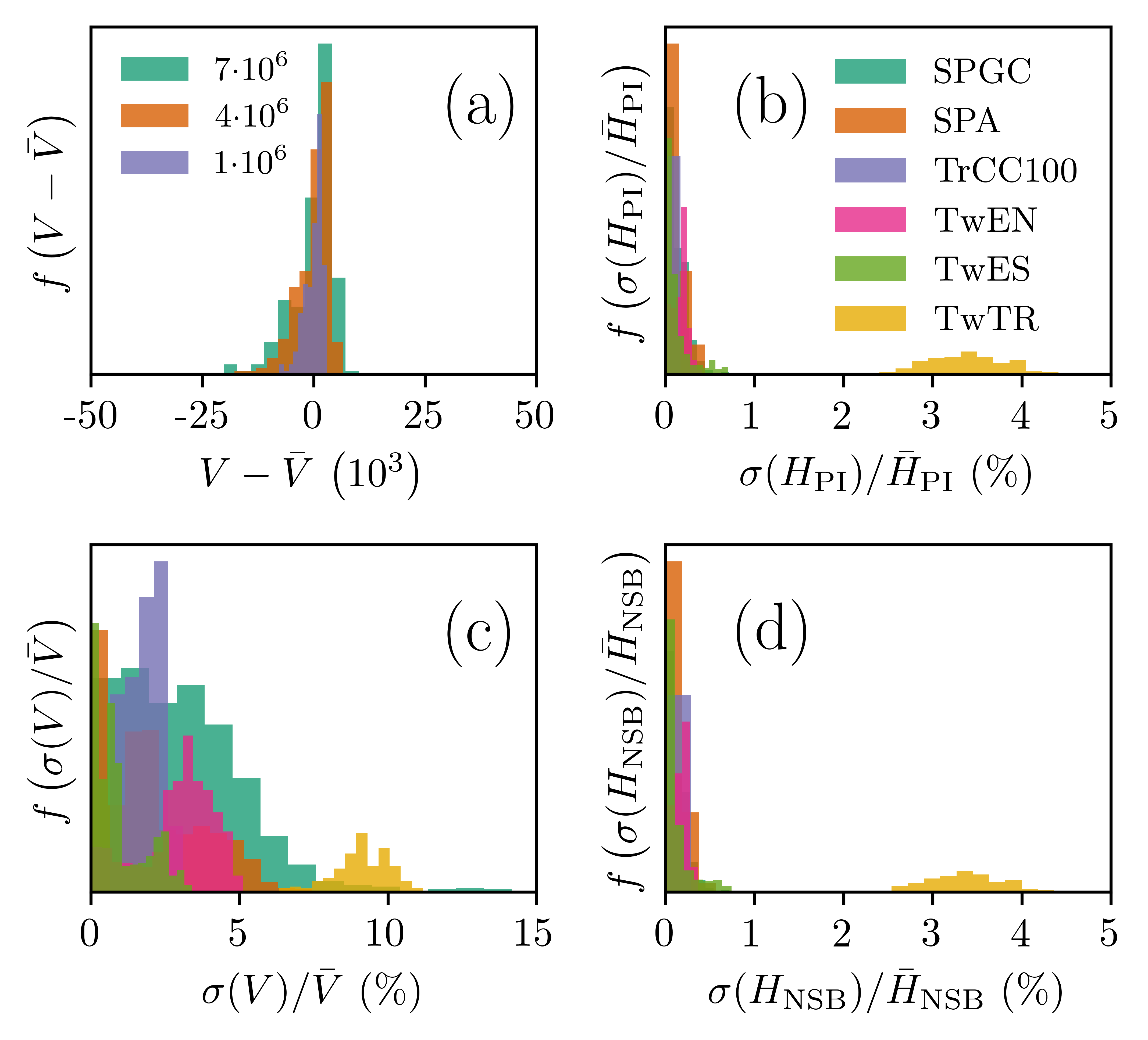}
  \caption{(a) Typical probability distribution of the measurements of the vocabulary for a fixed length $l$ indicated in the figure for the TwEs dataset, whose standard deviation $\sigma\left( V \right) $ accounts for around 5\% of the average $\bar{V}$, as seen in (c). It is thus reasonable to characterize the distribution with its mean value. Similar results are obtained for the distribution of entropies, both using the plug-in (b) and the NSB (d) estimators.}
  \label{fig:relsigma}
\end{figure}

In Fig.~\ref{fig:relsigma}(a) we show the shape of a typical distribution for an arbitrary value of $l$. To measure the different magnitudes corresponding to the length $l$, we characterize the distribution of Fig.~\ref{fig:relsigma}(a) by its mean value and its standard deviation. We believe that this is an appropriate approach because, as shown in Fig.~\ref{fig:relsigma}(b), Fig.~\ref{fig:relsigma}(c) and Fig.~\ref{fig:relsigma}(d), the vast majority of standard deviations fall below 5\% of the mean value of the distribution.

\section{Contribution to the entropy of the second regime of the frequency distribution}
\label{app:2zipfcont}

The contribution to the entropy of the ranks in the second regime of the Zipf distribution can be checked using several numerical ways.

One option is to use ratio of frequencies as a proxy to determine the contribution to the entropy of the words ranked beyond the critical value, i.e., the rank at which the Zipf exponent changes from around 1 to approximately 2. As shown in Ref.~\cite{Altmann2017gen}, the contribution of the ranks below the critical value ($b$ in Ref.~\cite{Altmann2017gen}, $r_c$ in Table~\ref{tab:rh}) is around 80\%.

Another option is to directly compute
\begin{align}
    \notag
    R_H &= \frac{\sum\limits_{r_i = r_c}^V p_{r_i} \ln p_{r_i}}{\sum\limits_{r_i=1}^{V} p_{r_i} \ln p_{r_i}} \\
    & = \frac{\sum\limits_{r_i = r_c}^V p_{r_i} \ln p_{r_i}}{\sum\limits_{r_i=1}^{r_c-1} p_{r_i} \ln p_{r_i} +\sum\limits_{r_i = r_c}^V p_{r_i} \ln p_{r_i}},
\end{align}
which constitutes the numerical ratio between the entropy of the ranks above the critical value and the total entropy of a given corpus of vocabulary $V$, $p_{r_i}$ being the normalized frequency of the word with rank $r_i$, as in Eq.~\eqref{eq:incr_entropy}. To compute $r_c$, we find the crossing point between the fits of the two Zipf regimes in Fig.~\ref{fig:zhl}(left). We provide the values of $R_H$ for the six corpora under study in Table~\ref{tab:rh}.

A more theoretical option would be to consider the fit to Zipf law instead of the empirical values of the frequencies. Let the distribution of frequencies be described by
\begin{equation}
f =
\begin{cases} 
c_1 r^{-a_1} & \text{if } r < r_c, \\
c_2 r^{-a_2} & \text{if } r \geq r_c.
\end{cases}
\end{equation}

We can obtain $c_1$ and $c_2$ imposing continuity and normalizing the frequency distribution, so that
\begin{align}
    c_1 &= L \left[\sum\limits_{r_i=1}^{r_c-1} r_i^{-a_1} +\sum\limits_{r_i=r_c}^{V} r_i^{-a_2} \right]^{-1}, \\
    c_2 &= L \left[r_c^{a_1-a_2} \sum\limits_{r_i=1}^{r_c-1} r_i^{-a_1} +\sum\limits_{r_i=r_c}^{V} r_i^{-a_2} \right]^{-1}.
\end{align}

In this sense, we can compute
\begin{widetext}
\begin{equation}
    R'_H = \frac{c_2 \sum\limits_{r_i=r_c}^{V} r_i^{-a_2} \ln \left[ \frac{c_2}{L} r_i^{-a_2} \right]}{c_1 \sum\limits_{r_i=1}^{r_c-1} r_i^{-a_1} \ln \left[ \frac{c_1}{L} r_i^{-a_1} \right] + c_2 \sum\limits_{r_i=r_c}^{V} r_i^{-a_2} \ln \left[ \frac{c_2}{L} r_i^{-a_2} \right]}
\end{equation}
\end{widetext}
using the text length, $L$, and vocabulary size, $V$, of each corpus. The results are shown in Table~\ref{tab:rh}.

\begin{table}[t] 
\caption{$R_H$, $R'_H$, Zipf exponents $a_1$ and $a_2$, and Zipf critical point $r_c$, for the six different corpora under study.\label{tab:rh}}
\begin{tabular}{cccccc}
\hline \hline
Corpus  & $R_H$ & $R'_H$ & $a_1$ & $a_2$ & $r_c$ \\ 
\hline
SPGC    & 0.17  & 0.13   & 1.12  & 1.86  & 7947  \\ 
SPA     & 0.15  & 0.12   & 1.10  & 1.90  & 17,197 \\ 
TRCC100 & 0.23  & 0.21   & 0.91  & 1.83  & 28,744 \\ 
TWEN    & 0.19  & 0.12   & 1.19  & 1.79  & 4896  \\ 
TWES    & 0.14  & 0.13   & 1.10  & 1.87  & 15,057 \\ 
TWTR    & 0.24  & 0.20   & 0.92  & 1.80   & 58,040 \\ 
\hline \hline
\end{tabular}
\end{table}

As seen in Table~\ref{tab:rh}, the low  values of both metrics $R_H$ and $R'_H$ support the assumption that the first regime of the distribution (the one closest to an exact Zipf law) contributes mainly to the entropy; specifically, around 80\%.

\section{Range selection for the fits}
\label{app:range}

\begin{figure}[t]
  \centering
  \includegraphics[width=0.95\linewidth]{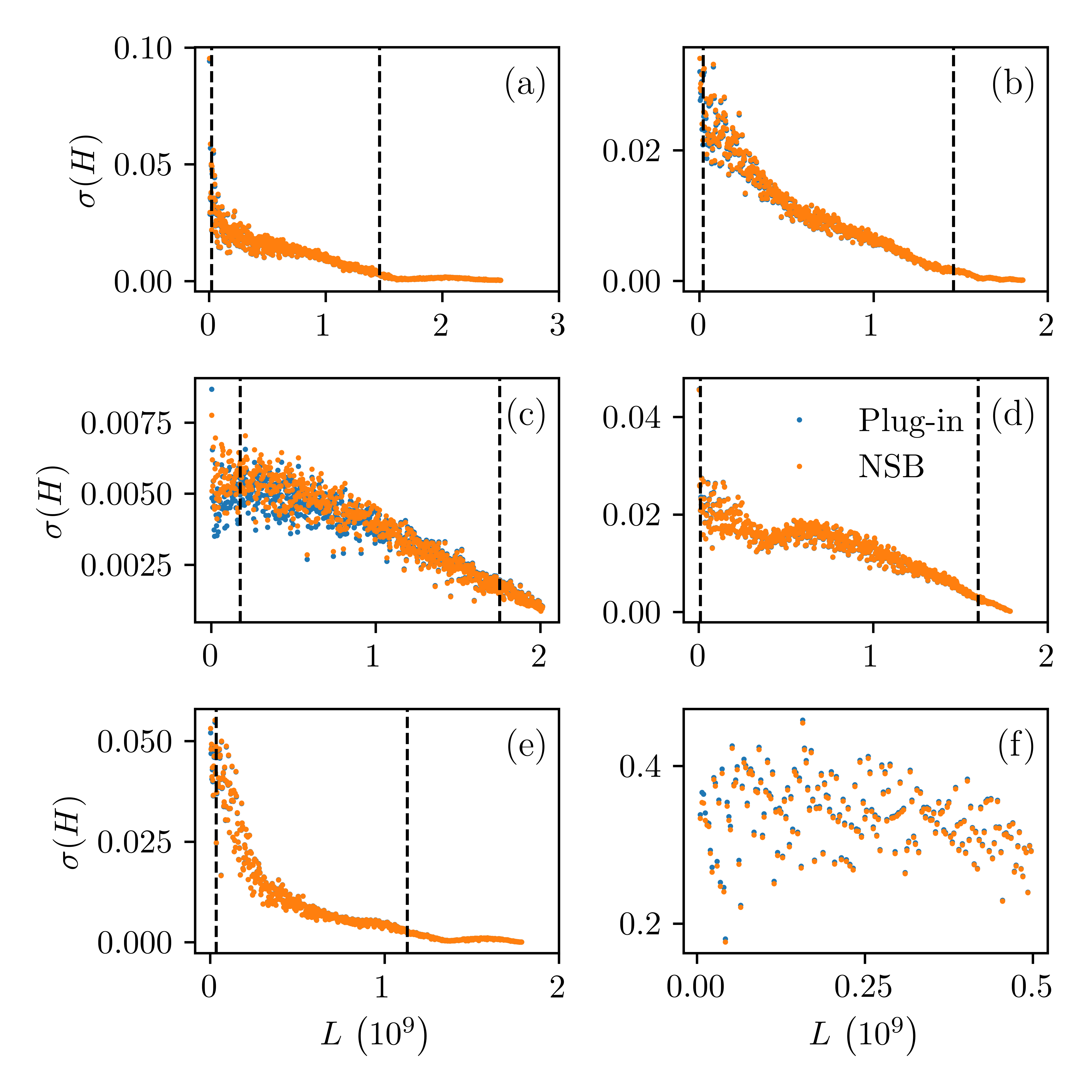}
  \caption{Range selection for the fits in Fig.~\ref{fig:h_ttr} for the SPGC (a), SPA (b), TRCC100 (c), TwEN (d), TwES (e), and TwTR (f) corpora studying the standard deviation for the entropy distributions $\sigma(H)$ in terms of text length $L$. In vertical dashed lines we show the selected interval for the fit. We depict the obtained values for both the plug-in estimator (orange dots) and the NSB estimator (blue dots). As seen, we do not see significant differences between both estimators.}
  \label{fig:ranges_all}
\end{figure}

As stated in Section~\ref{sec:entropy}, to select the appropriate range in which to perform the fit of Eq.~\ref{eq:fith} we search the values in which the data that we are fitting is not affected either by the noisy small $V$ regime or by the artificially low $\sigma_H$ observed when the sample size approaches the corpus size. For example, the range for SPA corpus is $0.0025 < \sigma_H < 0.025$. This heuristic $\sigma_H$ condition works well for all the corpora, as seen in Fig.~\ref{fig:ranges_all}, with the exception of the Turkish ones. In both Turkish corpora, $\sigma_H$ is systematically higher than in the Spanish and English corpora, and has a different shape. Hence, in the case of Turkish corpora we choose a range of high vocabularies far from the beginning and the end of the sampling range for the TRCC100, corresponding to 249,370$< V <$5,500,902, and all the available data for the TwTR.

\begin{table}[H] 
\centering
\caption{Results of the fit of Eq.~\eqref{eq:fith} to the corpora shown in Fig.~\ref{fig:h_nsb_ttr} using the NSB estimator for the computation of the entropy. As evaluations of the goodness of the fit, we show the coefficient of determination, $\rho^2$, the Spearman correlation coefficient, $\rho_\mathrm{s}$, and the distance correlation, $\rho_\mathrm{d}$. In all cases, $p\text{-value}<0.0001$.\label{tab:fithv_nsb}}
\begin{tabular}{cccccc}
\hline \hline
Corpus & $p_1$ & $p_2$ & $\rho^2$ & $\rho_\mathrm{s}$ & $\rho_\mathrm{d}$ \\
\hline
SPGC    & 0.03  & 6.87  & 0.90  & 0.93  & 0.90 \\
SPA     & 0.05  & 6.90  & 0.96  & 0.98  & 0.93 \\
TRCC100 & 0.01  & 9.50  & 0.85  & 0.84  & 0.82     \\
TwEN    & 0.04  & 6.95  & 0.93  & 0.94  & 0.90 \\
TwES    & 0.04  & 6.98  & 0.88  & 0.96  & 0.90 \\
TwTR    & 0.23  & 8.05  & 0.50  & 0.71  & 0.68     \\
\hline \hline
\end{tabular}
\end{table}

\section{Measurements using NSB estimator}
\label{app:nsb}

In this Appendix we show the results using the Nemenman–Shafee–Bialek (NSB) estimator~\cite{nemenman2001entropy} for the word entropy of the texts.

\begin{figure}[H]
  \centering
  \includegraphics[width=0.95\linewidth]{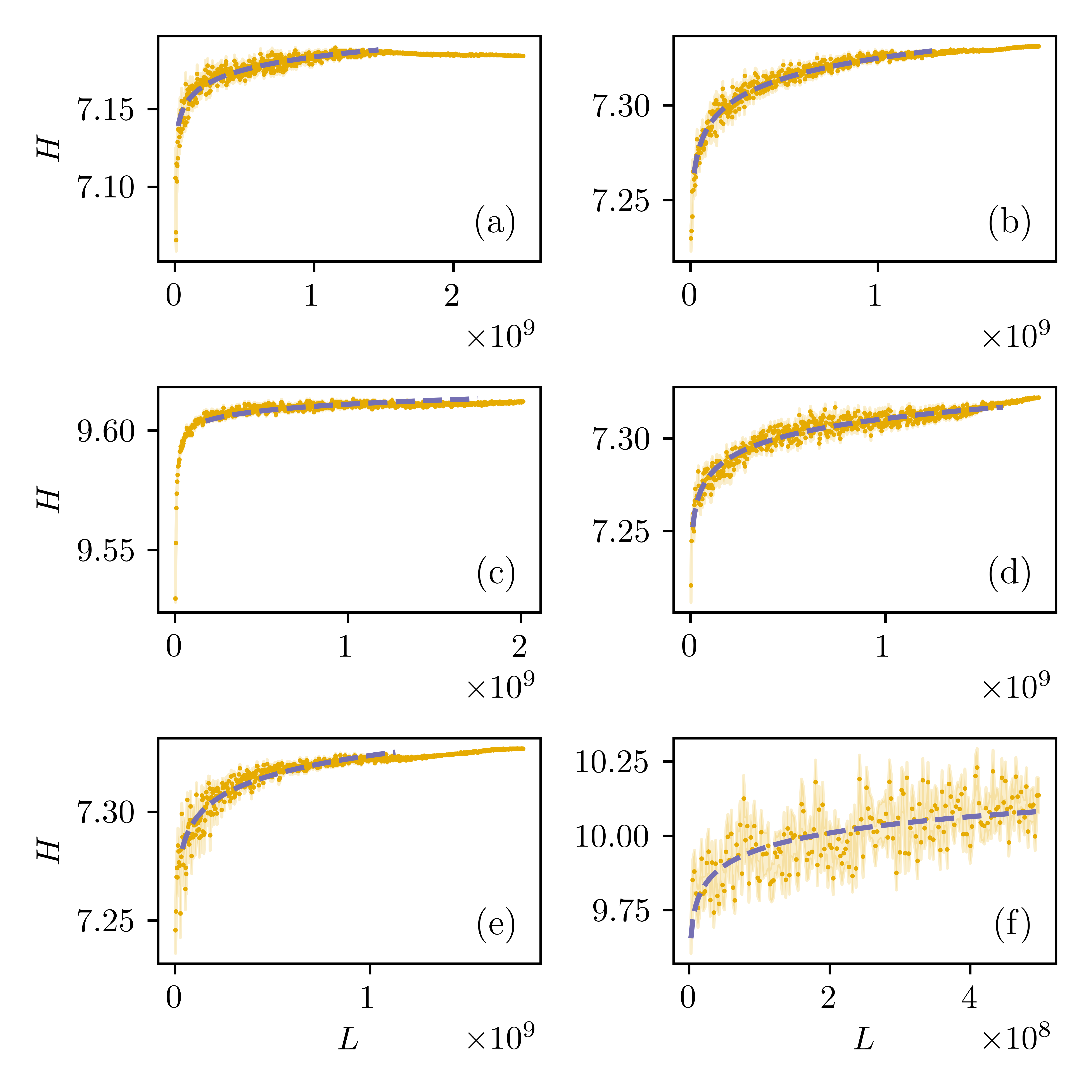}
  \caption{Measured entropy $H$ using the NSB estimator in terms of text length $L$ for the SPGC (a), SPA (b), TRCC100 (c), TwEN (d), TwES (e), and TwTR (f) corpora. The parameters of the fit and its goodness evaluation are shown in Table~\ref{tab:fithv_nsb}.}
  \label{fig:h_nsb_ttr}
\end{figure}

\begin{figure}[H]
  \centering
  \includegraphics[width=0.95\linewidth]{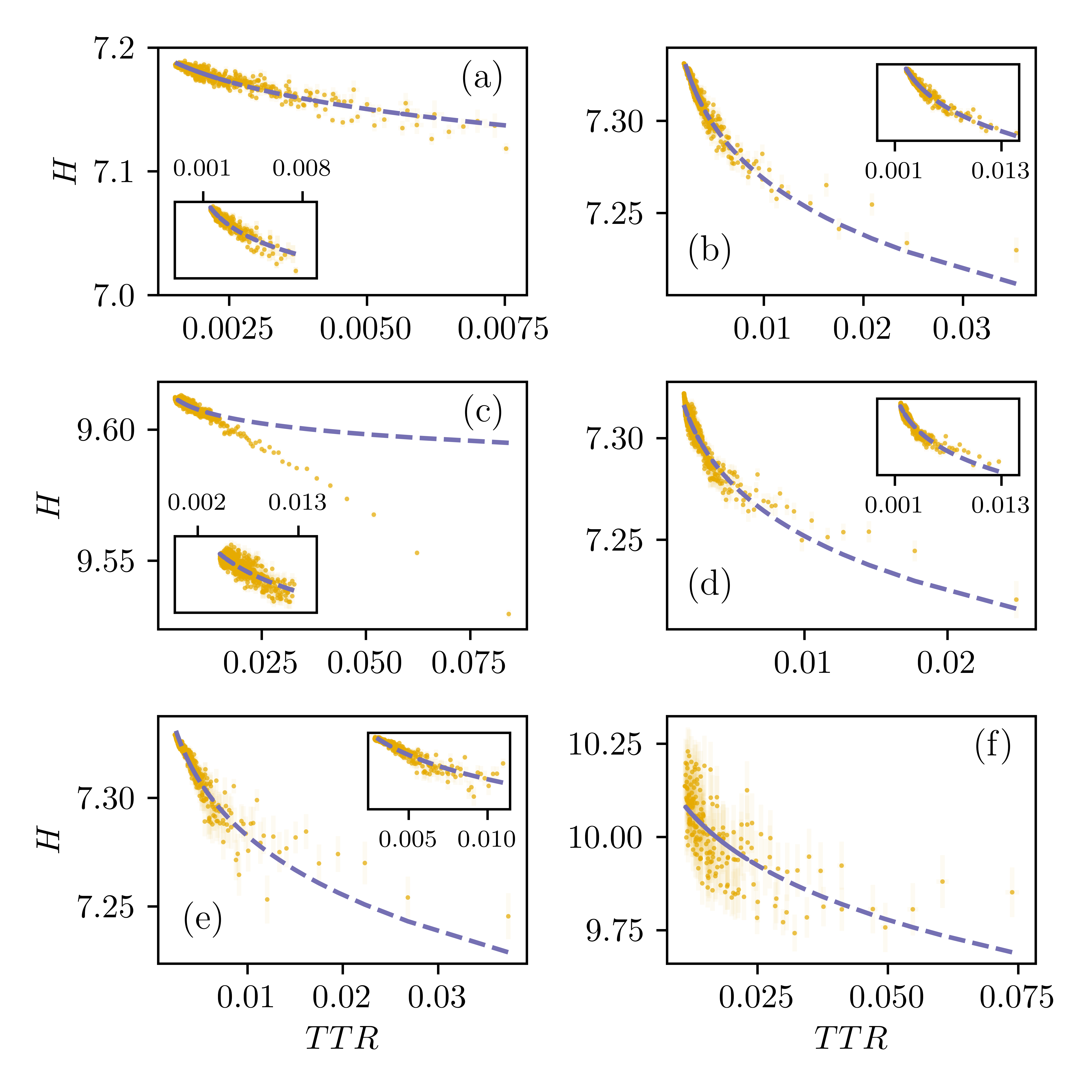}
  \caption{$H$ in terms of $TTR$ for the SPGC (a), SPA (b), TRCC100 (c), TwEN (d), TwES (e), and TwTR (f) corpora for $L > 2.5\times 10^6$ tokens using the NSB estimator. With a dashed line we plot the fit to Eq.~\eqref{eq:h_ttr}, whose goodness is assessed in Table~\ref{tab:fit_h_ttr_nsb}. Inside the insets we show the data and the fit in the confidence interval, as exemplified in Fig.~\ref{fig:range_selection}.}
  \label{fig:h_ttr_fit_nsb}
\end{figure}

The differences between the results of the PI estimator and the NSB estimator are relatively small, as can be seen by comparing Fig.~\ref{fig:h_nsb_ttr} with Fig.~\ref{fig:h_ttr}(left) and Fig.~\ref{fig:h_ttr_fit_nsb} with Fig.~\ref{fig:h_ttr_fit}. In general, the coefficient of determination tends to be slightly lower with the NSB estimator (see Tables~\ref{tab:fithv_nsb} and \ref{tab:fit_h_ttr_nsb}). However, the other metrics used to evaluate the goodness of fit show favorable outcomes overall. Since the NSB estimator produces results consistent with those of the PI estimator, this agreement lends credibility to the PI results.

\begin{table}[H]
\centering
\caption{Evaluation of the goodness of the fit of Eq.~\eqref{eq:h_ttr} for the NSB estimator of $H$. We provide the values of the coefficient of determination, $\rho^2$, the Spearman correlation coefficient, $\rho_\mathrm{s}$, and the distance correlation, $\rho_\mathrm{d}$. In all cases, $p\text{-value}<0.0001$.\label{tab:fit_h_ttr_nsb}}
\begin{tabular}{cccccc}
\hline \hline
Corpus & $p_3$ & $p_4$ & $\rho^2$ & $\rho_\mathrm{s}$ & $\rho_\mathrm{d}$ \\
\hline
SPGC    & 0.03 & 6.96 & 0.89  & -1.00 & 0.93 \\
SPA     & 0.52 & 5.88 & 0.96  & -1.00 & 0.96 \\
TRCC100 & 0.08 & 9.40 & 0.69  & -1.00 & 1.00 \\
TwEN    & 0.44 & 6.10 & 0.92  & -1.00 & 0.99 \\
TwES    & 0.36 & 6.32 & 0.89  & -1.00 & 1.00 \\
TwTR    & 2.11 & 4.10 & 0.51  & -1.00 & 0.98 \\
\hline \hline
\end{tabular}
\end{table}

\bibliography{biblio}

\end{document}